# Tiny Machine Learning for Real-Time Aquaculture Monitoring: A Case Study in Morocco

Achraf Hsain
School of Science and Engineering
Al Akhawayn University
Ifrane, Morocco
A.Hsain@aui.ma

Yahya Zaki
School of Science and Engineering
Al Akhawayn University
Ifrane, Morocco
Y.Zaki@aui.ma

Othman abaakil
School of Science and Engineering
Al Akhawayn University
Ifrane, Morocco
O.Abaakil@aui.ma

Hibat-allah Bekkar
School of Science and Engineering
Al Akhawayn University
Ifrane, Morocco
H.Bekkar@aui.ma

Yousra Chtouki
School of Science and Engineering
Al Akhawayn University
Ifrane, Morocco
Y.Chtouki@aui.ma

*Abstract*:

*Aquaculture, the farming of aquatic organisms, is a rapidly growing industry facing challenges such as water quality fluctuations, disease outbreaks, and inefficient feed management. Traditional monitoring methods often rely on manual labor and are time consuming, leading to potential delays in addressing issues. This paper proposes the integration of low-power edge devices using Tiny Machine Learning (TinyML) into aquaculture systems to enable real-time automated monitoring and control, such as collecting data and triggering alarms, and reducing labor requirements. The system provides real-time data on the required parameters such as pH levels, temperature, dissolved oxygen, and ammonia levels to control water quality, nutrient levels, and environmental conditions enabling better maintenance, efficient resource utilization, and optimal management of the enclosed aquaculture space. The system enables alerts in case of anomaly detection. The data collected by the sensors over time can serve for important decision-making regarding optimizing water treatment processes, feed distribution, feed pattern analysis and improve feed efficiency, reducing operational costs. This research explores the feasibility of developing TinyML-based solutions for aquaculture monitoring, considering factors such as sensor selection, algorithm design, hardware constraints, and ethical considerations. By demonstrating the potential benefits of TinyML in aquaculture, our aim is to contribute to the development of more sustainable and efficient farming practices.*

## I. INTRODUCTION

The National pisciculture center in Azrou, Morocco, manages its aquaponics system through traditional manual methods, which requires a large amount of human labor and often prone to inaccuracies. By integrating technology, we anticipate significant improvements in sustainability and productivity. The smart monitoring system offers display of real-time data, automated alerts and suggesting
Solutions to given problems, giving precise control over the aquaculture environment, reducing the need for manual intervention, and enhancing the overall operational efficiency. This goal aligns with the center's goals of advancing sustainable aquaculture practices, contributing to environmental conservation, and promoting innovative agricultural techniques. The research will not only benefit the National Pisciculture Center in Azrou but also build on existing advancement to further modernize aquaculture globally through smart technology. Agriculture is one important sector in Morocco, contributing to approximately 14% to the nation's Gross Domestic Product (GDP) [1]. However, Morocco faces significant climatic challenges: water scarcity, temperature variation, and ecological disruption, which constitute a threat to the farmers. Recently, water minister Nizar Baraka insisted on the severity that Morocco had reached since 2024 is the sixth consecutive year of drought, leading to 67% reduction in rainfall if compared to the usual seasonal average. While in 2022 the water reserves were filled to 33%, they have worsened to be filled at only 23.5% of capacity in 2024 [2]. Aquaponics offers a promising alternative; it is a sustainable farming system that combines aquaculture with hydroponics in a closed-loop water recirculation system. Fish residues are converted by bacteria into nutrients for plants, with the water then filtered and returned to the fish tanks. The National pisciculture center in Azrou, Morocco, manages its aquaponics system through traditional manual methods, which requires a large amount of human labor and often prone to inaccuracies. The system uses 90% less water than traditional agriculture, is highly productive, pesticide-free, and provides a sustainable solution for food production [3]. The center of Azrou has recognized its potential and had been using it. However, the system is quite complex and needs constant monitoring from farmers. To address this issue, we propose smart monitoring of the aquaculture fish farming space using TinyML to enable real-time monitoring and data analysis.
TinyML is revolutionizing real-time aquaculture monitoring by integrating advanced technologies to improve fish farming practices. This approach takes advantage of IoT, machine learning, and innovative sensor systems to optimize water quality and fish health. Real-time Monitoring Systems IoT Integration: Systems such as the one developed by [4]. utilize Arduino microcontrollers to monitor pH, turbidity, and water levels, sending alerts for abnormal conditions via SMS and web interfaces. Water Quality Management: The AquaBot system [4] automates water quality monitoring and recommends suitable fish species based on real-time data, achieving 94 Advanced Imaging Techniques. Microscopic tracking: [5] adapted Time-of- Flight cameras with IR lasers for tracking fish larvae, ensuring minimal disturbance while providing precise monitoring in varying water conditions [5]. As for Fish Detection and Tracking: [6] demonstrated the effectiveness of AI in automating fish monitoring, using deep learning models to enhance detection accuracy in aquaculture [6].

Although TinyML offers significant advances in aquaculture monitoring, challenges remain to ensure the affordability and accessibility of these technologies for small-scale farmers, which could limit widespread adoption. The remainder of the paper is organized as follows: II Background and literature review III. Challenges: ethical considerations, IV The methodology. V Limitations and Future Work,
VI. CONCLUSION

## II. BACKGROUND AND LITERATURE REVIEW

*A. Existing Monitoring Methods in Aquaculture.*

Fish farming is rapidly gaining popularity worldwide, necessitating close monitoring of the environment to ensure optimal growth conditions, prevent disease outbreaks, and maximize production efficiency. The Centre National d'Hydrobiologie et de Pisciculture Azrou reports that traditional monitoring methods in intensive aquaculture systems primarily rely on direct observation and simple instrumentation [7]. These methods include measuring parameters such as temperature, pH, dissolved oxygen, and salinity using portable sensors. Data collection is typically carried out at intervals ranging from hours to days, which may result in adverse conditions going undetected during these time gaps [8].

Advanced monitoring systems, however, have begun to utilize wireless sensor networks (WSNs) to continuously monitor critical parameters such as water quality and feeding behavior [9]. These systems provide real-time data, allowing farmers to take timely actions [10]. Despite their advantages, WSNs in aquaculture are often associated with high power consumption, costly maintenance, and expensive setup and operational costs. As such, these systems may be out of reach for many small-scale farmers. Furthermore, the harsh aquatic environment poses constant challenges to the durability of sensors and the efficiency of data transmission Previous Applications of Tiny ML in Agriculture and Related Fields [10] [11].

Tiny ML, with its ability to perform intelligent data processing on edge devices without the need for continuous data transfer to centralized servers, has garnered significant interest across multiple industries, including agriculture [12]. TinyML has been successfully applied to early crop disease detection, determining optimal irrigation times, and predicting pest infestations. For instance, sensor nodes equipped with TinyML models have been deployed to monitor soil moisture levels and alert farmers only when irrigation is necessary, thus conserving water [11]. In pest management, TinyML has been used to distinguish between insect species by analyzing their acoustic signals, enabling timely interventions [14]

Beyond agriculture, TinyML has been applied in environmental monitoring, particularly in measuring air and water quality [15]. Microcontroller-based systems using TinyML have been installed in urban environments to detect pollution levels, providing instant feedback to residents and governmental authorities [16]. Additionally, TinyML has been employed in remote sensing for disaster management and environmental conservation, offering real-time analysis of critical data [16] [17].

These applications demonstrate the potential of TinyML to improve the efficiency and sustainability of various practices by enabling local, real-time decision-making.

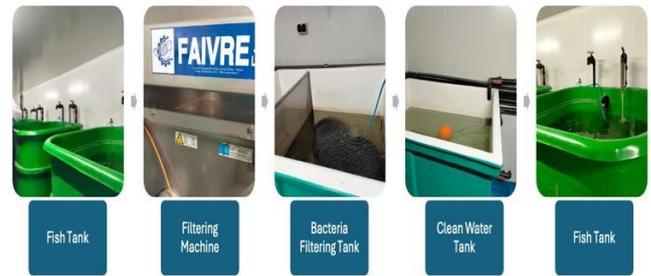

**Figure 1 : the fish farming room in Azrou Pisciculture**

The water of the fish tank goes through the filtering machine to remove all the particles, then it goes through the bacteria filtering machine to remove micro particles, then clean water with proper temperature is dispensed to the fish tanks consecutively. The room contains 8 to 10 fish tanks. Once the fish matures enough to survive in outdoor ecosystems, it is removed from the fish tanks and redistributed in natural environment, both in natural and artificial rivers and lakes in the region. The goal is to preserve species in the environment.

## III. CHALLENGES: ETHICAL CONSIDERATION

*A. Privacy and Surveillance:*

Real-time monitoring in aquaculture, particularly with video systems or constant data collection, introduces potential privacy risks. While these systems primarily capture environmental data such as water quality there's a possibility, they might unintentionally record information about individuals working near or on fish farms. Without appropriate safeguards in place, such data could lead to unauthorized surveillance, raising significant privacy concerns [17].

In emerging sectors such as aquaculture in Morocco, adherence to privacy regulations is crucial. Frameworks such as the General Data Protection Regulation (GDPR) or local Moroccan laws must be followed to ensure data isn't misused. Moreover, open and transparent communication with farmers and communities is key. Building trust can prevent any perception of unwarranted surveillance [18].

*B. Data Integrity and Accuracy:*

Deploying TinyML in a dynamic aquaculture environment presents challenges in data accuracy. Environmental factors can introduce noise into sensor readings, which complicates analysis and, if not addressed, may lead to incorrect conclusions about water quality or fish health. Such errors can have significant implications, potentially harming both the aquaculture sector and the surrounding ecosystems [19]. To ensure decisions based on this data are sound, it's essential to employ robust validation techniques. This ensures data integrity, particularly when influencing critical decision-making processes that affect farmers' livelihoods and aquatic sustainability [20]. Ethical data handling—focused on responsible collection, storage, and use—is essential to avoid any manipulation or exploitation.

*C. Dual-Use Potential and Misuse of Technology:*

Though TinyML offers numerous benefits for real-time monitoring in aquaculture, it carries risks. The same technologies designed to enhance aquaculture can be repurposed for harmful activities. For example, monitoring systems could be exploited for unauthorized fishing or even environmental damage. Additionally, data collected could be manipulated, posing threats to conservation and local economies [21].

Addressing these risks requires stringent security protocols. Limiting access to sensitive data, aligning its use with ethical standards, and ongoing oversight from authorities can prevent misuse. These measures will ensure the technology supports sustainable aquaculture rather than undermining it.

*D. Promoting Environmental Sustainability with Technology:*

Morocco's TinyML-driven aquaculture effort is focused on developing a sustainable, environmentally friendly sector. Aquaculture operators can maintain optimal fish farming conditions, limit the danger of disease outbreaks, and optimize water usage by utilizing real-time data. TinyML systems offer accurate monitoring and control, resulting in effective use of natural resources.

The proper application of this technology aims to contribute to broader environmental goals. Sustainable development is not just an objective; it's a necessity. Through smart monitoring, aquaculture operations can coexist with and even support surrounding ecosystems. This balance promotes both economic growth and environmental responsibility, turning Morocco's aquaculture industry into a model for sustainable development globally.

## IV. METHODOLOGY

To enhance the monitoring and management of the aquaculture system at the National Pisciculture Center in Azrou, we propose a smart monitoring and management system. This system is divided into three primary components: Input (Sensors), Microcontrollers, and Output (Data Analysis and Alerts).

*A. Input: Sensor Selection and Data Acquisition:*

A variety of advanced sensors were selected to monitor critical water and air parameters essential for maintaining optimal conditions within the aquaculture system. The sensors were chosen based on their precision, reliability, and ease of integration with microcontrollers. *a) Water Parameters:*

- **pH:** The Atlas Scientific EZO-pH kit was selected for its high precision and continuous monitoring capability.
- **Temperature:** The DS18B20 Waterproof Temperature Sensor was chosen for its accuracy and ease of integration.
- **Dissolved Oxygen (DO):** The Atlas Scientific EZO-DO kit is crucial for maintaining adequate oxygen levels for both fish and plants.
- **Ammonia, Nitrite, and Nitrate ($NH_3$/$NO_2$/$NO_3$):** Vernier Ammonium Ion-Selective Electrode and Vernier Nitrite Ion-Selective Electrode were selected to monitor toxic levels, ensuring they remain within safe limits.
- **Total Dissolved Solids (TDS):** The DFRobot Gravity Analog TDS Sensor is used to assess water clarity and overall quality.
- **Turbidity:** The Aqua TROLL Turbidity Sensor is proposed for monitoring water clarity.

*b) Model Development and Selection:*

In previous research [22], multiple models were developed and trained, including Convolutional Neural Networks (CNN), Deep Long Short-Term Memory (LSTM), Gated Recurrent Units (GRU), and Artificial Neural Networks (ANN), for forecasting key metrics in aquaponics systems.

The CNN model demonstrated superior performance, achieving the lowest error in all metrics: MAE, RMSE, MSE, MdAPE

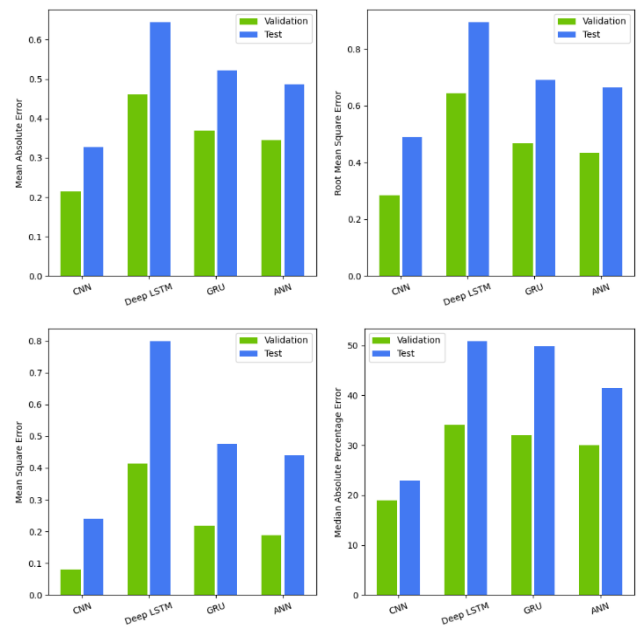

**Figure2: Comparison of metric values for different Deep Learning Architectures for the forecast**

We opted to use the **Median Absolute Percentage Error (MdAPE)** instead of the conventional **Mean Absolute Percentage Error (MAPE)** to account for the high number of outliers and abrupt variations in metrics, such as turbidity, which are often caused by erroneous measurements. These anomalies can significantly distort the representation of the model's true performance. Additionally, since data collection will continue after the physical implementation is deployed, we anticipate that the model's performance will improve over time. For the current study, the CNN model will be fine-tuned using aquaculture-specific data, which is less abundant in online repositories. This fine-tuning will include additional metrics not previously considered in the aquaponics context, setting the stage for continuous model improvement as real-time data is collected and processed from the aquaculture facility.

| created_at | entry_id | temperature | turbidity | dissolved_oxg | ph | ammonia | nitrate | population | fish_length | fish_weight |
|---|---|---|---|---|---|---|---|---|---|---|
| 2021-06-19 00:00:05 | 1889 | 24.875 | 100 | 4.505 | 8.43365 | 0.45842 | | 193 | 50 | 7.11 | 2.91 |
| 2021-06-19 00:01:02 | 1890 | 24.875 | 100 | 6.601 | 8.43818 | 0.45842 | | 194 | 50 | 7.11 | 2.91 |
| 2021-06-19 00:01:22 | 1891 | 24.875 | 100 | 15.797 | 8.42457 | 0.45842 | | 192 | 50 | 7.11 | 2.91 |
| 2021-06-19 00:01:44 | 1892 | 24.9375 | 100 | 5.046 | 8.43365 | 0.45842 | | 193 | 50 | 7.11 | 2.91 |
| 2021-06-19 00:02:07 | 1893 | 24.9375 | 100 | 38.407 | 8.40641 | 0.45842 | | 192 | 50 | 7.11 | 2.91 |
| 2021-06-19 00:02:27 | 1894 | 24.9375 | 100 | 3.862 | 8.42003 | 0.45842 | | 193 | 50 | 7.11 | 2.91 |
| 2021-06-19 00:02:47 | 1895 | 24.875 | 100 | 2.831 | 8.43818 | 0.45842 | | 194 | 50 | 7.11 | 2.91 |
| 2021-06-19 00:03:07 | 1896 | 24.9375 | 100 | 5.012 | 8.42911 | 0.45842 | | 193 | 50 | 7.11 | 2.91 |
| 2021-06-19 00:03:27 | 1897 | 24.9375 | 100 | 2.916 | 8.42911 | 0.45842 | | 192 | 50 | 7.11 | 2.91 |
| 2021-06-19 00:03:47 | 1898 | 24.875 | 100 | 17.005 | 8.43365 | 0.45842 | | 193 | 50 | 7.11 | 2.91 |
| 2021-06-19 00:04:31 | 1900 | 24.875 | 100 | 6.964 | 8.48358 | 0.45842 | | 191 | 50 | 7.11 | 2.91 |
| 2021-06-19 00:05:11 | 1902 | 24.9375 | 100 | 3.465 | 8.42911 | 0.45842 | | 187 | 50 | 7.11 | 2.91 |

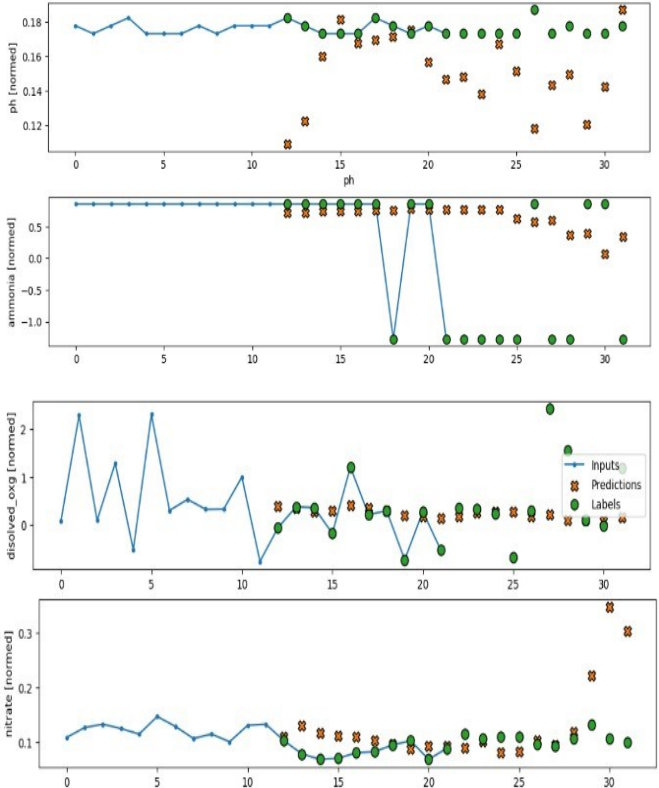

**Figure3: Snapshot of data used for the pretraining and evaluation of the models**

**Figure3: Examples of forecast results produced by the CNN architecture**

**Measurement Frequency:** In our previous aquaponics research, the TinyML model was implemented to make predictions every 10 minutes using historical data from the last 30 minutes [22]. For this aquaculture use case, we estimate that running the model every hour will suffice.

*B. Microcontrollers: Data Collection and Transmission:*

An Arduino microcontroller was selected due to its compatibility with the chosen sensors and ease of integration. The microcontroller will collect data from all sensors and transmit it to a central processing unit for further analysis. This setup ensures consistent real-time data collection, enabling proactive management of the aquaculture system.

*C. Output: Data Analysis and Predictive Monitoring:*

The collected data will be processed using the TinyML algorithm designed for real-time monitoring and predictive analysis. The system will then activate the corresponding motors for any parameters deviating from optimal ranges and provide maintenance suggestions. This approach minimizes the need for manual intervention, offering precise control over the aquaculture environment and enhancing overall operational efficiency.

The system will then take the necessary actions based on whether the predicted values fall under metrics optimal ranges below [23]:

| Metric | Optimal Range |
|---|---|
| Temperature | 25 °C – 32 °C |
| PH | 6.5 – 8.5 |
| Dissolved Oxygen (DO) | >5 mg/L |
| Total Dissolved Solids (TDS) | 400 mg/L |
| Nitrite (NO2) | <0.2 |
| Nitrate (NO3) | 0-100 |
| Turbidity | 30 – 80 cm |

**Table 1: Optimal Range of Metric Values for Aquaculture System**

One of the key actions the system will perform is sending notifications to staff if any forecasted metric values fall outside the determined optimal range. These notifications will specify the exact metric causing the issue, enabling staff to take appropriate corrective actions promptly.

Additionally, the system will incorporate a safety measure, allowing the staff at the Azrou Center to quickly shut down the system in case of a malfunction. This feature is crucial for preventing potential harm to the fish or water quality, ensuring the overall safety of the aquaculture environment.

## V. LIMITATIONS AND FUTURE WORK:

One of the primary limitations of this study is the lack of available data, both online and onsite, for the aquaculture environment. To address this, we utilized pre-trained models based on an aquaponics dataset [24], as there are significant similarities between aquaponics and aquaculture systems. This approach provides a foundational model that can be continuously fine-tuned after deployment in the aquaculture environment as more relevant data becomes available. In the future, the plan is to integrate computer vision technologies into the system, allowing cameras to identify and collect fish-related information. This will enable the system to provide more detailed and useful insights to the staff.

Additionally, the aim is to implement a physical prototype of the system for testing and data collection at the National Pisciculture Center in Azrou. Feedback from this implementation will be used to make further improvements to the system. Another interesting research direction we plan to explore is leveraging the collected data for analytics to identify any valuable insights, such as data related to fish feeding or fish density. These insights could potentially help us further improve the system.

## VI. CONCLUSION:

In this paper, we presented a framework for a TinyML-based edge system designed to monitor the aquaculture environment and improve operational efficiency at the National Pisciculture Center in Azrou. We proposed an architecture aimed at reducing the workload of staff by automating the monitoring of critical metrics through AI driven analysis. Our approach includes a pre-trained machine learning algorithm, which can be deployed on an edge device to forecast key metrics on an hourly basis. When anomalies are detected, the system will send detailed notifications to the staff, ensuring timely interventions.

Additionally, we incorporated safety measures to responsibly safeguard the fish and water in the event of a system malfunction. Looking ahead, we outlined our future plans, including the integration of computer vision technologies for more advanced fish monitoring, and the physical implementation and testing of the system to gather feedback and further refine its functionality.